\documentclass[a4paper,UKenglish,cleveref, autoref, thm-restate]{lipics-v2021}

\pdfoutput=1 
\hideLIPIcs  
\nolinenumbers

\usepackage{booktabs}
\usepackage[justification=centering]{caption}
\usepackage{listings}
\usepackage{multicol}
\usepackage{syntax}

\usepackage{agda}
\usepackage[utf8]{inputenc}
\usepackage{newunicodechar}

\newunicodechar{𝕋}{\ensuremath{\mathbb{T}}}
\newunicodechar{ℚ}{\ensuremath{\mathbb{Q}}}
\newunicodechar{ℤ}{\ensuremath{\mathbb{Z}}}
\newunicodechar{ℕ}{\ensuremath{\mathbb{N}}}
\newunicodechar{ℓ}{\ensuremath{\ell}}
\newunicodechar{∞}{\ensuremath{\infty}}

\newunicodechar{∣}{\ensuremath{\mid}}
\newunicodechar{⊔}{\ensuremath{\sqcup}}
\newunicodechar{∸}{\ensuremath{\dotdiv}}
\newunicodechar{∥}{\ensuremath{\parallel}}

\newunicodechar{≤}{\ensuremath{\leq}}
\newunicodechar{≱}{\ensuremath{\ngeq}}
\newunicodechar{▷}{\ensuremath{\vartriangleright}}
\newunicodechar{⇿}{\ensuremath{\leftrightarrow}} 
\newunicodechar{∷}{\ensuremath{::}}

\newunicodechar{⌊}{\ensuremath{\lfloor}}
\newunicodechar{⌋}{\ensuremath{\rfloor}}
\newunicodechar{∨}{\ensuremath{\vee}}

\newunicodechar{≢}{\ensuremath{\nequiv}}
\newunicodechar{≈}{\ensuremath{\approx}}
\newunicodechar{≉}{\ensuremath{\napprox}}
\newunicodechar{≟}{\ensuremath{\stackrel{?}{=}}}

\newunicodechar{∀}{\ensuremath{\forall}}
\newunicodechar{∃}{\ensuremath{\exists}}
\newunicodechar{⇒}{\ensuremath{\Rightarrow}}
\newunicodechar{∧}{\ensuremath{\wedge}}
\newunicodechar{∈}{\ensuremath{\in}}
\newunicodechar{∉}{\ensuremath{\notin}}
\newunicodechar{∴}{\ensuremath{\therefore}}
\newunicodechar{∎}{\ensuremath{\qed}}
\newunicodechar{∘}{\ensuremath{\circ}}
\newunicodechar{⊕}{\ensuremath{\oplus}}

\newunicodechar{ₐ}{\ensuremath{_a}}
\newunicodechar{ₑ}{\ensuremath{_e}}
\newunicodechar{ᵢ}{\ensuremath{_i}}
\newunicodechar{ₖ}{\ensuremath{_k}}
\newunicodechar{ₗ}{\ensuremath{_l}}
\newunicodechar{ₘ}{\ensuremath{_m}}
\newunicodechar{ₙ}{\ensuremath{_n}}
\newunicodechar{ₓ}{\ensuremath{_x}}

\newunicodechar{₀}{\ensuremath{_0}}
\newunicodechar{₁}{\ensuremath{_1}}
\newunicodechar{₂}{\ensuremath{_2}}
\newunicodechar{₃}{\ensuremath{_3}}
\newunicodechar{₊}{\ensuremath{_+}}

\newunicodechar{ᵉ}{\ensuremath{^e}}
\newunicodechar{ᵐ}{\ensuremath{^m}}
\newunicodechar{ʳ}{\ensuremath{^r}}

\newunicodechar{∅}{\ensuremath{\emptyset}}
\newunicodechar{⊥}{\ensuremath{\bot}}

\newcommand{\vehicle}[1]{{\texttt{#1}}}

\graphicspath{{./figures/}}

\title{Vehicle: Interfacing Neural Network Verifiers with Interactive Theorem Provers}

\titlerunning{Vehicle: NNs to ITPs} 

\author{Matthew L. Daggitt\footnote{Corresponding author}}{Department of Computer Science, Heriot-Watt University, Edinburgh, UK \and \url{http://www.myhomepage.edu} }{johnqpublic@dummyuni.org}{https://orcid.org/0000-0002-1825-0097}{}
\author{Wen Kokke}{Mathematically Structured Programming Group, University of Strathclyde, Glasgow, UK}{joanrpublic@dummycollege.org}{[orcid]}{}
\author{Robert Atkey}{Mathematically Structured Programming Group, University of Strathclyde, Glasgow, UK}{joanrpublic@dummycollege.org}{[orcid]}{}
\author{Luca Arnaboldi}{Laboratory for Foundations of Computer Science, University of Edinburgh, Edinburgh, UK}{joanrpublic@dummycollege.org}{[orcid]}{}
\author{Ekaterina Komendantskya}{Department of Computer Science, Heriot-Watt University, Edinburgh, UK}{joanrpublic@dummycollege.org}{[orcid]}{}

\authorrunning{M.L. Daggitt, W. Kokke, B. Atkey, L. Arnaboldi, E. Komendantskya}

\Copyright{Matthew L. Daggitt, Wen Kokke, Bob Atkey, Luca Arnaboldi, Ekaterina Komendantskya}



\category{} 

\relatedversion{} 

\funding{This work was funded by the AISEC grant under EPSRC numbers EP/T026952/1, EP/T026960/1, and EP/T027037/1.}

\acknowledgements{We would like to thank the Marabou development team for their support and advice with integrating Vehicle with Marabou.}

\EventEditors{John Q. Open and Joan R. Access}
\EventNoEds{2}
\EventLongTitle{13th International Conference on Interactive Theorem Proving (ITP 2022)}
\EventShortTitle{ITP 2022}
\EventAcronym{ITP}
\EventYear{2022}
\EventDate{December 24--27, 2016}
\EventLocation{Little Whinging, United Kingdom}
\EventLogo{}
\SeriesVolume{42}
\ArticleNo{23}

\begin{document}

\maketitle

\begin{abstract}
Verification of neural networks is currently a hot topic in automated theorem proving. 
Progress has been rapid and there are now a wide range of tools available that can verify properties of networks with hundreds of thousands of nodes. 
In theory this opens the door to the verification of larger control systems that make use of neural network components. 
However, although work has managed to incorporate the results of these verifiers to prove larger properties of individual systems, there is currently no general methodology for bridging the gap between verifiers and interactive theorem provers (ITPs).

In this paper we present Vehicle, our solution to this problem. 
Vehicle is equipped with an expressive domain specific language for stating neural network specifications which can be compiled to both verifiers and ITPs. 
It overcomes previous issues with maintainability and scalability in similar ITP formalisations by using a standard ONNX file as the single canonical representation of the network. 
We demonstrate its utility by using it to connect the neural network verifier Marabou to Agda and then formally verifying that a car steered by a neural network never leaves the road, even in the face of an unpredictable cross wind and imperfect sensors. 
The network has over 20,000 nodes, and therefore this proof represents an improvement of 3 orders of magnitude over prior proofs about neural network enhanced systems in ITPs.
\end{abstract}

\section{Introduction}
\label{sec:introduction}

In the last decade deep neural networks have made their way into systems used in everyday life and, as with any system that interacts directly with humans, it is highly desirable to have formal guarantees about their behaviour. However, these systems present a challenge for the verification community as typically the neural networks are used in domains where a formal specification of the desired behaviour remains elusive. Furthermore, their size and inability to be decomposed into components that tackle identifiable sub-tasks mean that they are usually viewed as black-box components, which makes traditional verification difficult to apply.

Nonetheless, the discovery of adversarial attacks on neural networks in 2014~\cite{szegedy2013intriguing}, has spurred the Automated Theorem Proving (ATP) community to develop neural network verifiers. These are  based on SMT solvers~\cite{ehlers2017formal, katz2019marabou}, abstract interpretation~\cite{singh2019abstract} or interval bounded arithmetic~\cite{xiang2018output} and aim to prove linear (or occasionally semi-definite) relationships between the inputs and the outputs of the network. Progress has been rapid and they are now reaching the point where they are powerful enough to verify properties of networks with tens or even hundreds of thousands of nodes.

The majority of the ATP community's attention has been focused on using them to prove the robustness of the networks against adversarial attack, and there has been comparatively little work on using them to prove the functional correctness of larger systems that incorporate neural network components. Although there are several reasons for this, including the difficulty in coming up with a specification for the neural network in the first place, we believe that a key missing component on the technical side is that no one has yet integrated them with interactive theorem provers (ITPs).

As when incorporating other ATP tools such as SMT solvers into ITPs~\cite{blanchette2013extending, ekici2017smtcoq}, one must translate high-level statements in the ITP into low level queries for the verifier. However, there are several additional challenges unique to the integration of neural network verifiers:
\begin{enumerate}
\item \textbf{Modelling mismatch} - in an ITP a neural network is usually modelled as a function which can be composed with a larger system. However, neural network verifiers model a network as a relation between its inputs and outputs. As discussed in Sections~\ref{sec:verifiers}~\&~\ref{sec:functional-to-relational}, the translation from the former to the latter is not straightforward.

\item \textbf{Scalability} - modern networks can contain millions or even billions of nodes, whereas most ITPs will consume excessive amounts of memory when representing even very small networks. For example, recent formalisations in Coq~\cite{bagnall2019certifying, de2022use} have have worked with networks of just 10 or 20 nodes.

\item \textbf{Maintainability} - most neural networks are not static artefacts, and regularly undergo further training as new data is collected. For obvious practical reasons, a formally verified representation of the network within an ITP is unlikely to be the canonical representation deployed in a production system. Instead, specialised file formats are used to distribute and deploy networks. This raises the problem of how to maintain the faithfulness of the model (and the proofs) within the ITP with the rapidly evolving implementation stored elsewhere.

\item \textbf{Performance} - even with domain specific verifiers, verification of large neural networks can be exceedingly expensive, often taking hours or even days to complete. This has the potential to be very disruptive in ITPs whose workflow encourages users to regularly recheck the validity of their proof during development.

\item \textbf{Integration into other parts of the  neural network lifecycle} - verification is only a small part of constructing a neural network with formal guarantees. When trained using traditional methods, a network is highly unlikely to satisfy the required specifications. Recent work has shown how specifications may be integrated into the training of the network~\cite{fischer2019dl2} and gradient-based counter-example search~\cite{madry2017towards}. Therefore ideally the specification should be usable in these other tools as well. However, current ITPs are a poor environment in which to perform these complex and computationally intensive operations.
\end{enumerate}

We believe that in order for verification of neural network-enhanced systems to achieve wide-spread adoption, all of these issues must be addressed. Our vision is as follows.
The specification for a neural network should be stated once, in a high-level, human-readable format.
This specification should then be automatically translated to work with the appropriate tools in the different stages in the life-cycle of the network.
Importantly, there should only ever be \emph{one} representation of the network, stored in a format usable by the mainstream machine learning community.
For performance and modularity reasons, we also argue that an appropriate level of abstraction should be maintained at each stage.
Concretely, the training and verification tools should be able to inspect the internal neural network structure.
However, when writing the specification and using it to prove properties of the larger system in an ITP, the view of the network as a black-box function should be maintained where possible.
Finally to maintain interactivity in the ITP, the checking of the proof of system correctness should be decoupled from the checking of the proof of the neural network specification, so that former does not automatically trigger the latter.

\subsection{Contributions}

In this paper we present \emph{Vehicle}, a tool that implements the interaction between network specifications, verifiers and ITPs proposed in our vision above. In particular:
\begin{enumerate}
\item Vehicle is equipped with a high-level, domain specific language (DSL) for writing neural network specifications.
The Vehicle compiler automatically translates these specifications down to low-level queries for neural network verifiers, and then subsequently to high-level ITP code.
The latter can be used as an interface upon which proofs about the larger systems can be constructed.

\item Instead of modelling the network directly inside the ITP, Vehicle side-steps the maintainability issue outlined above by using an Open Neural Network Exchange (ONNX)~\cite{onnx2022} file as the single canonical representation of the network. The ONNX format is a widely-supported, cross-platform, training-framework independent, representation of neural networks~\cite{onnx2022}. The Vehicle compiler extracts the necessary internal implementation details from the ONNX file, and uses hashing to maintain the integrity of the specification in the ITP with respect to the underlying ONNX file.

\item As Vehicle stores the neural network externally and uses specialised neural network verifiers, its performance is dependent on that of the underlying verifier rather than the ITP. This means that Vehicle can potentially be used to verify systems that use networks with hundreds of thousands of nodes.

\item  Vehicle maintains interactivity in the ITP when checking the proof, as the generated interface code calls back to Vehicle rather than directly calling the neural network verifier. Vehicle then checks its cache to ascertain the verification status of the specification, thereby preventing costly and unnecessary re-verification.

\item Vehicle maintains the abstraction of the neural network as a black-box function. When writing the specification the user is only required to provide the type of the function implemented by the network. Similarly in the ITP interface, the neural network is presented as a function that can be used, but not decomposed.
\end{enumerate} 

In our current implementation, we target the SMT-based neural network verifier Marabou~\cite{katz2019marabou} and the interactive theorem prover Agda~\cite{norell2008dependently}. The latter was chosen due to the authors' expertise in it, and we acknowledge it is better suited to modelling systems rather than extracting formally verified executable code. However, the integration with Agda is deliberately very lightweight, with a single file in the compiler defining the translation, and a single Agda file defining macros for calling back to Vehicle. It would therefore be relatively simple to extend support to other ITPs such as Coq.

\begin{figure}[t]
\centering
\includegraphics[width=.8\textwidth]{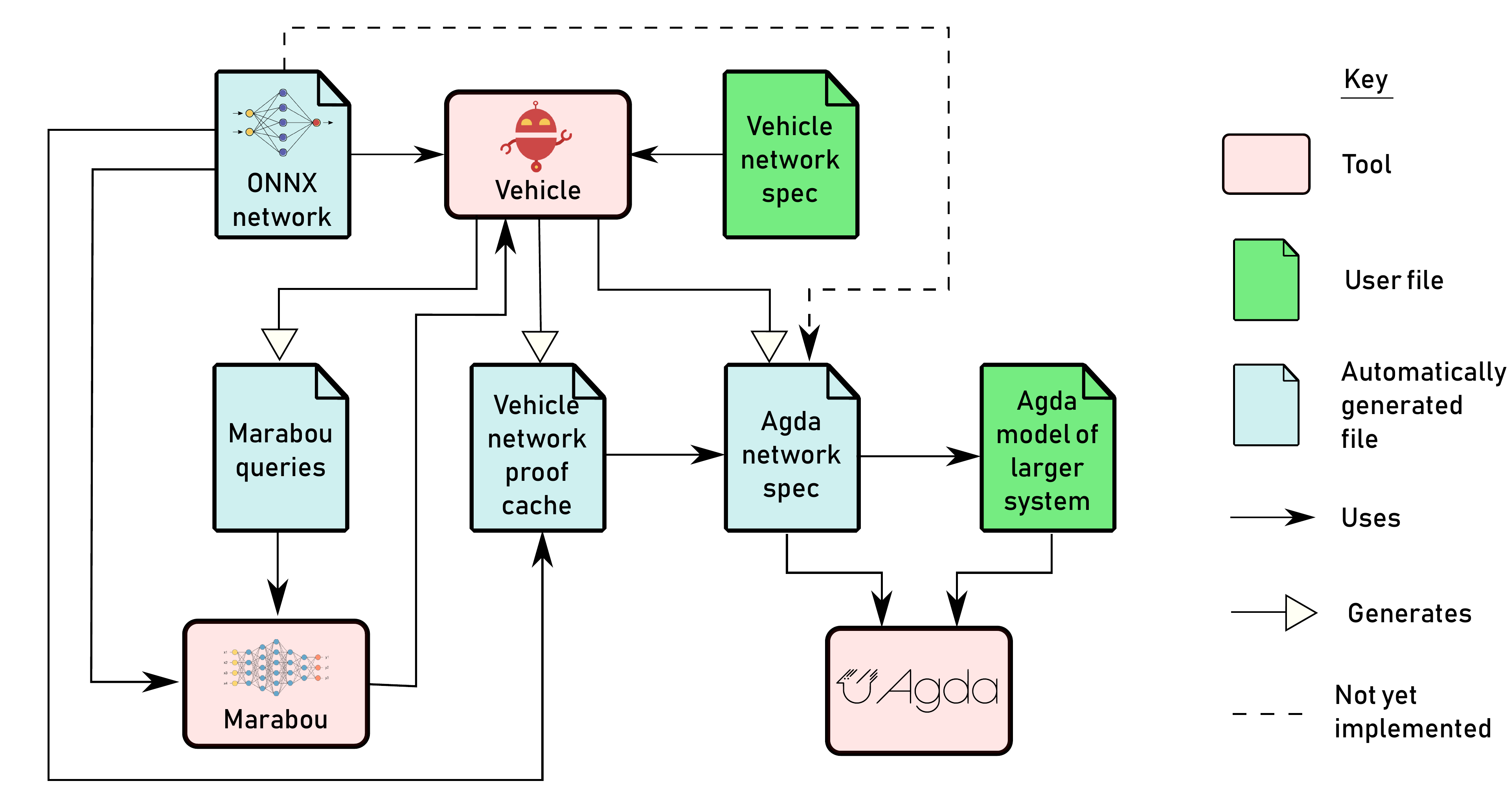}
\caption{The architecture of a Vehicle proof about a neural network enhanced system. The ability to evaluate the network directly in Agda has not yet been implemented. However, the integrity of the proof with respect to the network is maintained via the proof cache file, which also maintains the interactivity of the Agda file.}
\label{fig:proof-architecture}
\end{figure}

As an example of the utility of Vehicle, we use it to formally verify that a car controlled by a neural network will never leave the road, even in the presence of noisy sensor data and an unpredictable cross-wind. The neural network has over 20,000 nodes and therefore, as far as we aware, this proof represents an improvement of over 3 orders of magnitude when compared to previous proofs about neural network enhanced systems in ITPs. The architecture of the proof is shown in Figure~\ref{fig:proof-architecture}. All accompanying code, including Vehicle itself, is available online \cite{vehicle2022}.

Vehicle will also be of use to people who are not interested in integrating with ITPs. The existing interfaces to verifiers are very low-level, usually involving the user specifying a set of equalities or inequalities over the individual inputs and outputs of the neural network. As a large neural network typically can have thousands of inputs and outputs, creating such inequalities is both time consuming and error prone. Furthermore, the result is almost completely unintelligible to a non-technical domain expert. In contrast Vehicle can provide a much higher-level, human readable version of the specification.

The paper is laid out as follows. In Section~\ref{sec:background} we describe our running example problem and related work on verifiers, neural network verification efforts and other similar specification languages. In Section~\ref{sec:marabou-extension} we propose an extension to the query language for Marabou to allow it to support specifications involving multiple networks. In Section~\ref{sec:language}, we describe the Vehicle DSL and the novel passes in the Vehicle compiler used to generate verifier queries and the Agda code. Finally, in Section~\ref{sec:conclusions} we discuss future work.

\section{Background}
\label{sec:background}

\subsection{An example: staying on the road}
\label{sec:example}

\begin{figure}[ht]
\centering
\includegraphics[width=.5\textwidth]{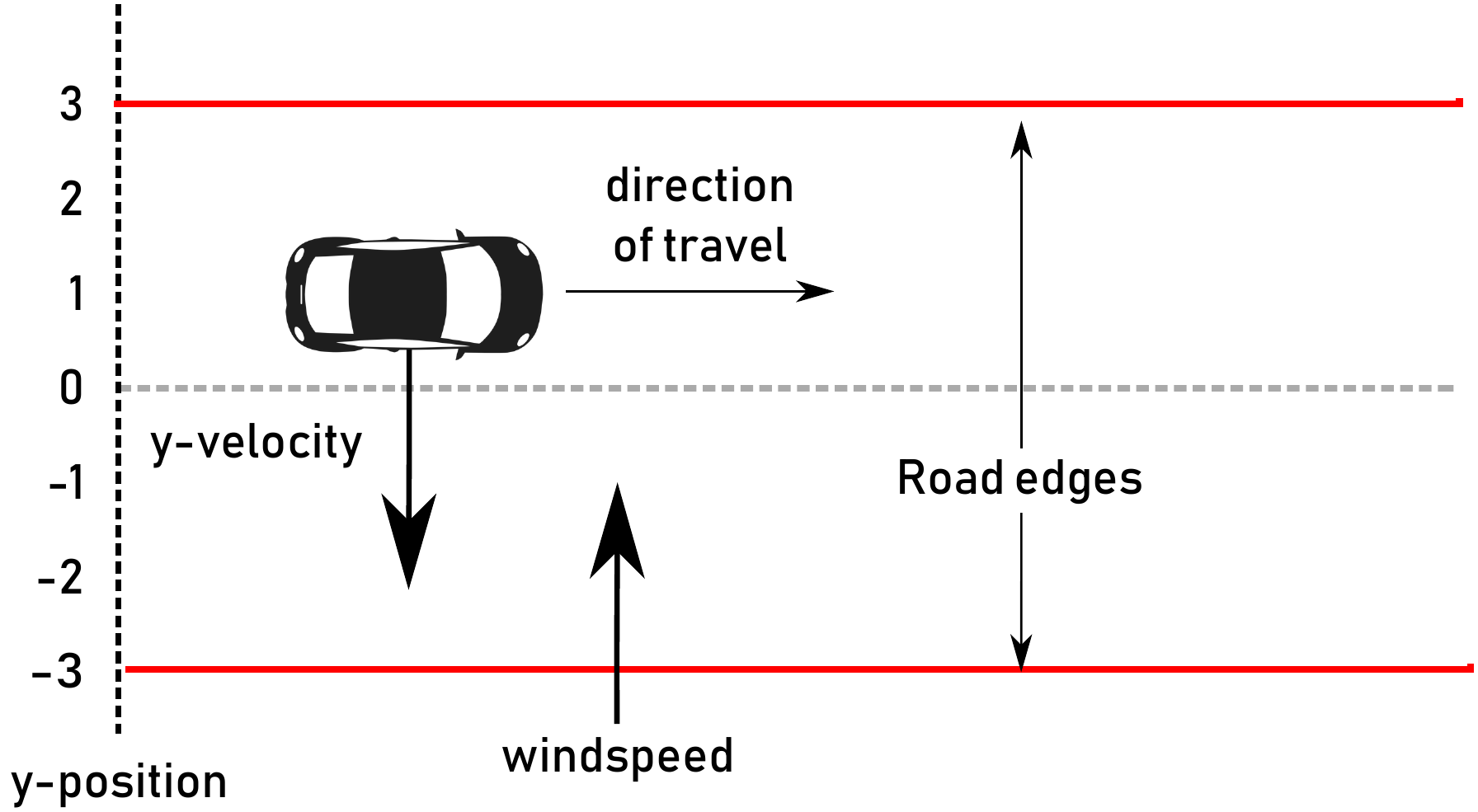}
\caption{A simple model of an autonomous vehicle compensating against a cross-wind.}
\label{fig:car-controller}
\end{figure}

We will use a modified version of the verification problem presented by Boyer, Green and Moore~\cite{boyer1990use} as a running example throughout this paper. In the scenario an autonomous vehicle is travelling along a straight road of width 6 parallel to the x-axis, with a varying cross-wind that blows perpendicular to the x-axis. The vehicle has an imperfect sensor that it can use to get a (possibly noisy) reading on its position on the y-axis, and can change its velocity on the y-axis in response. The car's controller takes in both the current sensor reading and the previous sensor reading and its goal is to keep the car on the road. The setup is illustrated in Figure~\ref{fig:car-controller}.

\noindent For simplicity, we assume that both the wind-speed and the car's velocity in the y-direction can grow arbitrarily large. As in \cite{boyer1990use} we discretise the model, and then formalise it in Agda as follows. The state of the system consists of the current wind speed, the position and velocity of the car and the most recent sensor reading. An oracle provides updates in the form of observations consisting of the shift in wind speed and the error on the sensor reading.
\begin{multicols}{2}
\begin{code}%
\>[0]\AgdaKeyword{record}\AgdaSpace{}%
\AgdaRecord{State}\AgdaSpace{}%
\AgdaSymbol{:}\AgdaSpace{}%
\AgdaPrimitive{Set}\AgdaSpace{}%
\AgdaKeyword{where}\<%
\\
\>[0][@{}l@{\AgdaIndent{0}}]%
\>[2]\AgdaKeyword{constructor}\AgdaSpace{}%
\AgdaInductiveConstructor{state}\<%
\\
\>[2]\AgdaKeyword{field}\<%
\\
\>[2][@{}l@{\AgdaIndent{0}}]%
\>[4]\AgdaField{windSpeed}\AgdaSpace{}%
\AgdaSymbol{:}\AgdaSpace{}%
\AgdaRecord{ℚ}\<%
\\
\>[4]\AgdaField{position}%
\>[14]\AgdaSymbol{:}\AgdaSpace{}%
\AgdaRecord{ℚ}\<%
\\
\>[4]\AgdaField{velocity}%
\>[14]\AgdaSymbol{:}\AgdaSpace{}%
\AgdaRecord{ℚ}\<%
\\
\>[4]\AgdaField{sensor}%
\>[14]\AgdaSymbol{:}\AgdaSpace{}%
\AgdaRecord{ℚ}\<%
\end{code}
\begin{code}%
\>[0]\AgdaKeyword{record}\AgdaSpace{}%
\AgdaRecord{Observation}\AgdaSpace{}%
\AgdaSymbol{:}\AgdaSpace{}%
\AgdaPrimitive{Set}\AgdaSpace{}%
\AgdaKeyword{where}\<%
\\
\>[0][@{}l@{\AgdaIndent{0}}]%
\>[2]\AgdaKeyword{constructor}\AgdaSpace{}%
\AgdaInductiveConstructor{observe}\<%
\\
\>[2]\AgdaKeyword{field}\<%
\\
\>[2][@{}l@{\AgdaIndent{0}}]%
\>[4]\AgdaField{windShift}%
\>[16]\AgdaSymbol{:}\AgdaSpace{}%
\AgdaRecord{ℚ}\<%
\\
\>[4]\AgdaField{sensorError}\AgdaSpace{}%
\AgdaSymbol{:}\AgdaSpace{}%
\AgdaRecord{ℚ}\<%
\end{code}
\end{multicols}
\noindent For the moment, we assume that we have some controller that takes in the current and the previous sensor reading and produces a resulting change in velocity:
\begin{code}%
\>[0]\AgdaFunction{controller}\AgdaSpace{}%
\AgdaSymbol{:}\AgdaSpace{}%
\AgdaRecord{ℚ}\AgdaSpace{}%
\AgdaSymbol{→}\AgdaSpace{}%
\AgdaRecord{ℚ}\AgdaSpace{}%
\AgdaSymbol{→}\AgdaSpace{}%
\AgdaRecord{ℚ}\<%
\end{code}
Given this, we can define the evolution of the system as follows:
\begin{code}%
\>[0]\AgdaFunction{initialState}\AgdaSpace{}%
\AgdaSymbol{:}\AgdaSpace{}%
\AgdaRecord{State}\<%
\\
\>[0]\AgdaFunction{initialState}\AgdaSpace{}%
\AgdaSymbol{=}\AgdaSpace{}%
\AgdaInductiveConstructor{state}\AgdaSpace{}%
\AgdaFunction{0ℚ}\AgdaSpace{}%
\AgdaFunction{0ℚ}\AgdaSpace{}%
\AgdaFunction{0ℚ}\AgdaSpace{}%
\AgdaFunction{0ℚ}\<%
\end{code}
\begin{code}%
\>[0]\AgdaFunction{nextState}\AgdaSpace{}%
\AgdaSymbol{:}\AgdaSpace{}%
\AgdaRecord{Observation}\AgdaSpace{}%
\AgdaSymbol{→}\AgdaSpace{}%
\AgdaRecord{State}\AgdaSpace{}%
\AgdaSymbol{→}\AgdaSpace{}%
\AgdaRecord{State}\<%
\\
\>[0]\AgdaFunction{nextState}\AgdaSpace{}%
\AgdaBound{o}\AgdaSpace{}%
\AgdaBound{s}\AgdaSpace{}%
\AgdaSymbol{=}\AgdaSpace{}%
\AgdaInductiveConstructor{state}\AgdaSpace{}%
\AgdaFunction{newWindSpeed}\AgdaSpace{}%
\AgdaFunction{newPosition}\AgdaSpace{}%
\AgdaFunction{newVelocity}\AgdaSpace{}%
\AgdaFunction{newSensor}\<%
\\
\>[0][@{}l@{\AgdaIndent{0}}]%
\>[2]\AgdaKeyword{where}\<%
\\
\>[2]\AgdaFunction{newWindSpeed}\AgdaSpace{}%
\AgdaSymbol{=}\AgdaSpace{}%
\AgdaField{windSpeed}\AgdaSpace{}%
\AgdaBound{s}\AgdaSpace{}%
\AgdaOperator{\AgdaFunction{+}}\AgdaSpace{}%
\AgdaField{windShift}\AgdaSpace{}%
\AgdaBound{o}\<%
\\
\>[2]\AgdaFunction{newPosition}%
\>[15]\AgdaSymbol{=}\AgdaSpace{}%
\AgdaField{position}\AgdaSpace{}%
\AgdaBound{s}%
\>[29]\AgdaOperator{\AgdaFunction{+}}\AgdaSpace{}%
\AgdaField{velocity}\AgdaSpace{}%
\AgdaBound{s}\AgdaSpace{}%
\AgdaOperator{\AgdaFunction{+}}\AgdaSpace{}%
\AgdaFunction{newWindSpeed}\<%
\\
\>[2]\AgdaFunction{newSensor}%
\>[15]\AgdaSymbol{=}\AgdaSpace{}%
\AgdaFunction{newPosition}\AgdaSpace{}%
\AgdaOperator{\AgdaFunction{+}}\AgdaSpace{}%
\AgdaField{sensorError}\AgdaSpace{}%
\AgdaBound{o}\<%
\\
\>[2]\AgdaFunction{newVelocity}%
\>[15]\AgdaSymbol{=}\AgdaSpace{}%
\AgdaField{velocity}\AgdaSpace{}%
\AgdaBound{s}%
\>[29]\AgdaOperator{\AgdaFunction{+}}\AgdaSpace{}%
\AgdaFunction{controller}\AgdaSpace{}%
\AgdaFunction{newSensor}\AgdaSpace{}%
\AgdaSymbol{(}\AgdaField{sensor}\AgdaSpace{}%
\AgdaBound{s}\AgdaSymbol{)}\<%
\end{code}
\begin{code}%
\>[0]\AgdaFunction{finalState}\AgdaSpace{}%
\AgdaSymbol{:}\AgdaSpace{}%
\AgdaDatatype{List}\AgdaSpace{}%
\AgdaRecord{Observation}\AgdaSpace{}%
\AgdaSymbol{→}\AgdaSpace{}%
\AgdaRecord{State}\<%
\\
\>[0]\AgdaFunction{finalState}\AgdaSpace{}%
\AgdaBound{xs}\AgdaSpace{}%
\AgdaSymbol{=}\AgdaSpace{}%
\AgdaFunction{foldr}\AgdaSpace{}%
\AgdaFunction{nextState}\AgdaSpace{}%
\AgdaFunction{initialState}\AgdaSpace{}%
\AgdaBound{xs}\<%
\end{code}
Given this setup we would like to prove the following property of the system:
\begin{theorem}
\label{thm:on-road}
Assuming that the wind-speed can shift by no more than 1 per unit time and that the sensor is never off by more than 0.25 then the car will never leave the road.
\end{theorem}

\noindent We define the pre-conditions of the theorem in Agda as follows:
\begin{code}%
\>[0]\AgdaFunction{ValidObservation}\AgdaSpace{}%
\AgdaSymbol{:}\AgdaSpace{}%
\AgdaRecord{Observation}\AgdaSpace{}%
\AgdaSymbol{→}\AgdaSpace{}%
\AgdaPrimitive{Set}\<%
\\
\>[0]\AgdaFunction{ValidObservation}\AgdaSpace{}%
\AgdaBound{o}%
\>[295I]\AgdaSymbol{=}\AgdaSpace{}%
\AgdaOperator{\AgdaFunction{∣}}\AgdaSpace{}%
\AgdaField{sensorError}\AgdaSpace{}%
\AgdaBound{o}\AgdaSpace{}%
\AgdaOperator{\AgdaFunction{∣}}\AgdaSpace{}%
\AgdaOperator{\AgdaDatatype{≤}}\AgdaSpace{}%
\AgdaFunction{0.25ℚ}\AgdaSpace{}%
\AgdaOperator{\AgdaFunction{×}}\AgdaSpace{}%
\AgdaOperator{\AgdaFunction{∣}}\AgdaSpace{}%
\AgdaField{windShift}\AgdaSpace{}%
\AgdaBound{o}\AgdaSpace{}%
\AgdaOperator{\AgdaFunction{∣}}\AgdaSpace{}%
\AgdaOperator{\AgdaDatatype{≤}}\AgdaSpace{}%
\AgdaFunction{1ℚ}\<%
\end{code}
and the post-condition as:
\begin{code}%
\>[0]\AgdaFunction{OnRoad}\AgdaSpace{}%
\AgdaSymbol{:}\AgdaSpace{}%
\AgdaRecord{State}\AgdaSpace{}%
\AgdaSymbol{→}\AgdaSpace{}%
\AgdaPrimitive{Set}\<%
\\
\>[0]\AgdaFunction{OnRoad}\AgdaSpace{}%
\AgdaBound{s}%
\>[295I]\AgdaSymbol{=}\AgdaSpace{}%
\AgdaOperator{\AgdaFunction{∣}}\AgdaSpace{}%
\AgdaField{position}\AgdaSpace{}%
\AgdaBound{s}\AgdaSpace{}%
\AgdaOperator{\AgdaFunction{∣}}\AgdaSpace{}%
\AgdaOperator{\AgdaDatatype{≤}}\AgdaSpace{}%
\AgdaFunction{3ℚ}\<%
\end{code}
which allows us to formalise the theorem as:
\begin{code}%
\>[0]\AgdaFunction{finalState-onRoad}\AgdaSpace{}%
\AgdaSymbol{:}\AgdaSpace{}%
\AgdaSymbol{∀}\AgdaSpace{}%
\AgdaBound{xs}\AgdaSpace{}%
\AgdaSymbol{→}\AgdaSpace{}%
\AgdaDatatype{All}\AgdaSpace{}%
\AgdaFunction{ValidObservation}\AgdaSpace{}%
\AgdaBound{xs}\AgdaSpace{}%
\AgdaSymbol{→}\AgdaSpace{}%
\AgdaFunction{OnRoad}\AgdaSpace{}%
\AgdaSymbol{(}\AgdaFunction{finalState}\AgdaSpace{}%
\AgdaBound{xs}\AgdaSymbol{)}\<%
\end{code}
As is standard when proving properties of large systems in a top-down manner, one eventually deduces the properties that must hold of the sub-components. In this case Theorem~\ref{thm:on-road} can be proved by formulating a suitable inductive hypothesis about the state of the system at each time-step. The full inductive proof can be found online~\cite{}, but the crucial part is that the inductive step requires the \AgdaFunction{controller} function to satisfy the following specification:
\begin{code}%
\>[0]\AgdaFunction{controller-lemma}\AgdaSpace{}%
\AgdaSymbol{:}%
\>[332I]\AgdaSymbol{∀}\AgdaSpace{}%
\AgdaBound{x}\AgdaSpace{}%
\AgdaBound{y}\AgdaSpace{}%
\AgdaSymbol{→}\AgdaSpace{}%
\AgdaOperator{\AgdaFunction{∣}}\AgdaSpace{}%
\AgdaBound{x}\AgdaSpace{}%
\AgdaOperator{\AgdaFunction{∣}}\AgdaSpace{}%
\AgdaOperator{\AgdaDatatype{≤}}\AgdaSpace{}%
\AgdaBound{3.25ℚ}\AgdaSpace{}%
\AgdaSymbol{→}\AgdaSpace{}%
\AgdaOperator{\AgdaFunction{∣}}\AgdaSpace{}%
\AgdaBound{y}\AgdaSpace{}%
\AgdaOperator{\AgdaFunction{∣}}\AgdaSpace{}%
\AgdaOperator{\AgdaDatatype{≤}}\AgdaSpace{}%
\AgdaBound{3.25ℚ}\AgdaSpace{}%
\AgdaSymbol{→}\AgdaSpace{}%
\AgdaOperator{\AgdaFunction{∣}}\AgdaSpace{}%
\AgdaFunction{controller}\AgdaSpace{}%
\AgdaBound{x}\AgdaSpace{}%
\AgdaBound{y}\AgdaSpace{}%
\AgdaOperator{\AgdaFunction{+}}\AgdaSpace{}%
\AgdaFunction{2ℚ}\AgdaSpace{}%
\AgdaOperator{\AgdaFunction{*}}\AgdaSpace{}%
\AgdaBound{x}\AgdaSpace{}%
\AgdaOperator{\AgdaFunction{-}}\AgdaSpace{}%
\AgdaBound{y}\AgdaSpace{}%
\AgdaOperator{\AgdaFunction{∣}}\AgdaSpace{}%
\AgdaOperator{\AgdaDatatype{<}}\AgdaSpace{}%
\AgdaBound{1.25ℚ}\AgdaSpace{}\<%
\end{code}
We implement the controller with a 3-layer densely connected neural network with over 20,000 nodes. The network is constructed in Tensorflow and all weights in the network are unique and non-zero.
The challenge is then to implement the Agda function \AgdaFunction{controller} with this neural network, and prove \AgdaFunction{controller-lemma}, without having to directly represent either the network or the proof directly in Agda.

 In the interests of full disclosure, the problem above as stated is hardly very challenging to solve for a neural network and the size of the network used is complete overkill. The scenario could be made more realistic by adding further, possibly conflicting, objectives for the controller to optimise for. For example, adding waypoints on the road that the car had to pass through, or regions it had to avoid. In such scenarios the statement of Theorem~\ref{thm:on-road} would remain the same, but the number of inputs to the controller would increase. We do not explore these more complicated scenarios in this paper due to space limitations. Nonetheless, the simple scenario presented above is sufficient to illustrate the abilities of Vehicle.

\subsection{Neural network verifiers}
\label{sec:verifiers}

Neural network verifiers can be roughly split into two families: those that are both sound and complete, like Marabou~\cite{katz2019marabou}, and those that are only sound, based on techniques such as abstract interpretation~\cite{singh2019abstract} or bounded interval arithmetic~\cite{xiang2018output}.  While the latter are more performant, their incompleteness means that when the verifier fails to show that the property holds, it is unknown whether there truly exists a counter-example.

There is no consensus on a single input format for verifications queries. Each solver defines its own, which makes interfacing with multiple solvers difficult. However, one commonality is that they implicitly model the neural network as a \emph{relation} between its inputs and outputs, where each of the network's inputs and outputs. For example when writing a Marabou query for a network with $m$ inputs and $n$ outputs, the input are labelled $x0$, $x1$, $\cdots$ $x[m-1]$ and the outputs are labelled $y0$, $y1$, $\cdots$ $y[n-1]$. Queries are then written as a series of inequalities involving these variables. For example, the Marabou queries required to verify \AgdaFunction{controller-lemma} are shown in Figure~\ref{fig:marabou-queries}.
\begin{figure}[h]
\begin{subfigure}[t]{0.45\textwidth}
\begin{lstlisting}[language=Haskell]
x0 >= -3.25
x0 <= 3.25
x1 >= -3.25
x1 <= 3.25
-y0 -2.0x0 +x1 >= 1.25
\end{lstlisting}
\caption{Query 1}
\end{subfigure}
\hfill
\begin{subfigure}[t]{0.45\textwidth}
\begin{lstlisting}[language=Haskell]
x0 >= -3.25
x0 <= 3.25
x1 >= -3.25
x1 <= 3.25
y0 2.0x0 -x1 >= 1.25
\end{lstlisting}
\caption{Query 2}
\end{subfigure}
\caption{Marabou queries for \AgdaFunction{controller-lemma}. The lemma is true iff Marabou cannot find an assignment of variables that satisfy the either set of inequalities.}
\label{fig:marabou-queries}
\end{figure}

In contrast to the solvers based on SMT technology, solvers based on abstract interpretation and interval arithmetic can only handle properties that involve reasoning about how regions in the input space get mapped to regions in the output space. Concretely this means that they are unable to solve inequalities that involve both input and output variables, such as the last line of the queries in Figure~\ref{fig:marabou-queries}. In order to maximise the number of interesting properties solvable, we therefore choose Marabou as the first solver to integrate into Vehicle.

\subsection{Verified neural network properties: the state of the art}

As discussed in the introduction, most of the work using neural network verifiers has focused on the verification of the robustness of the network. Informally, a network is robust if when you move no more than a small distance in the input space, then the result of the neural network should only move a small distance in the output space.
Several different types of robustness have been studied including, classification robustness, Lipschitz robustness, standard robustness and approximate classification robustness~\cite{casadio2021property}.

One of the first neural network verifiers, Reluplex~\cite{katz2017reluplex}, was also one of the first to verify non-trivial domain-specific properties, proving several results about ACAS Xu, a collision avoidance system for unmanned aircraft.
The ACAS Xu neural networks map inputs such as the aircraft's own speed and heading, and the angle and distance to the intruder aircraft to 5 different output actions, ranking each with a confidence score.
The action with the highest confidence score being the one that the system will perform.
In their verification the authors checked that the ACAS Xu worked as expected and avoided collisions, properties including, checking that a distant aircraft's path will mean that it remains clear of collision given various conditions such as angle and speed, checking that despite previous actions it will still perform the safe response given the new presence of a possible collision object etc.

More recent work in verification focused on the verification of a reinforcement learning based neural network controller~\cite{ivanov2019verisig}.
The authors show that they can verify that the function will always terminate with certain guarantees.
In their case study, the authors use an example of a car climbing up a hill and verify that the car will always reach the top of the hill with at least 90 reward.
This is the first paper of its kind formally verifying reachability properties for neural network components.
Building on this same techniques the authors expand their earlier work by formally verifying a reinforcement learning based controller for an autonomous vehicle~\cite{ivanov2020case}.
The authors first train some controllers for the vehicle and then verify its safety.
Specifically they are able to verify that the vehicle will never be less than 30cm away from a wall and consequently will not crash.
Unlike previous work by Katz~\cite{katz2017reluplex} which deals with ReLu activation functions, their tool Verisig supports neural networks with smooth activation functions (e.g. sigmoid), however it only scales to small networks of about 100 neurons.

\subsection{Other neural network specification languages}
\label{sec:other-property-languages}

Given the low-level input formats supported by the verifiers described in Section~\ref{sec:verifiers}, it is unsurprising there have been other attempts at coming up with a high-level property language. A comparison is shown in Table~\ref{tab:language-comparison}.

\NewDocumentCommand{\rot}{O{30} O{1em} m}{\makebox[#2][l]{\rotatebox{#1}{#3}}}%

\begin{table}[t]
\centering
\begin{tabular}{lccccccc}
\toprule
Language & 
Paradigm &
\rot{Host language} & 
\rot{Verifiers supported} & 
\rot{Typed} & 
\rot{ITP integration} & 
\rot{Multiple networks} & 
\rot{Probabilistic properties} \\
\midrule
Vehicle  & Functional & None & 1  & Yes & Yes & Yes & No \\
DNNV~\cite{shriver2021dnnv}     & Imperative & Python & 13 & No        & No  & No & No \\
Socrates~\cite{pham2020socrates} & Declarative & None & 2  & No        & No & No & Yes \\
\bottomrule
\end{tabular}
\caption{Comparison of existing property languages for neural networks}
\label{tab:language-comparison}
\end{table}

The first is the Deep Neural Network Verifier (DNNV) toolbox~\cite{shriver2021dnnv}, which has an internal Python DSL called DNNP. However its aims are somewhat orthogonal to that of Vehicle, as its primary focus is on providing a unified interface for many different verifiers. In particular, it has the ability to refactor the structure of the neural network to eliminate unsupported operators. However, DNNP is untyped and relies on Python semantics and therefore would be challenging to integrate into ITPs. Its dependence on Python also makes it difficult to use in other languages commonly used with neural networks such as C++.

The second, yet unpublished attempt, is Socrates~\cite{pham2020socrates}. Again it positions itself as a platform for neural network analysis, which aims to unify different tools. The DSL is comparatively limited, primarily supporting different forms of robustness properties using a structured JSON file. It also has the disadvantage that one must redefine the internal structure of the network within the specification. However, one notable feature is its ability to specify probabilistic queries, for example no more than 10\% of inputs violate the property.

\section{Multi-network specifications and the Marabou query language}
\label{sec:marabou-extension}

As described in Section~\ref{sec:verifiers}, Marabou queries use the variables $x_1, ..., x_m$ to represent the inputs to the network and $y_1, ..., y_n$ to represent the outputs of the network, and one consequence of this is that it is unable to represent queries that involve multiple networks or applying the same network to more than one input. This situation is suboptimal as there are several such queries that one might be interested in verifying. For example when using teacher-student training~\cite{gou2021knowledge}, one might want to prove that the output of the $student$ network is approximately equal to that of the $teacher$ network.
\begin{equation*}
\forall x : \: \mid student(x) - teacher(x) \mid \: \leq \epsilon
\end{equation*}
 Alternatively one might want to prove that a neural network $f$ is a monotonic function with respect to one or more of its inputs~\cite{wehenkel2019unconstrained}, which requires reasoning about the output of the network when applied to two distinct inputs.
\begin{equation*}
\forall x_1, x_2 : x_1 \leq x_2 \Rightarrow f (x_1) \leq f(x_2)
\end{equation*}

Luckily, the inability to solve such queries is a limitation of the query language rather than a fundamental shortcoming of the verification engine. This is because Marabou uses an SMT-based approach, and internally has no concept of a network. Instead it represents the nodes and edges simply as a set of variables and constraints between them. One of our purposes in designing Vehicle is to explore what a high-level interface for neural network verification should look like, and consequently we think there is significant value to be gained in targeting maximally expressive verifiers, even if they do not yet exist. We therefore now propose a conservative, backwards-compatible extension to the Marabou query language which will be targeted by the Vehicle compiler.

\begin{figure}[t]
\centering
\includegraphics[width=0.8\textwidth]{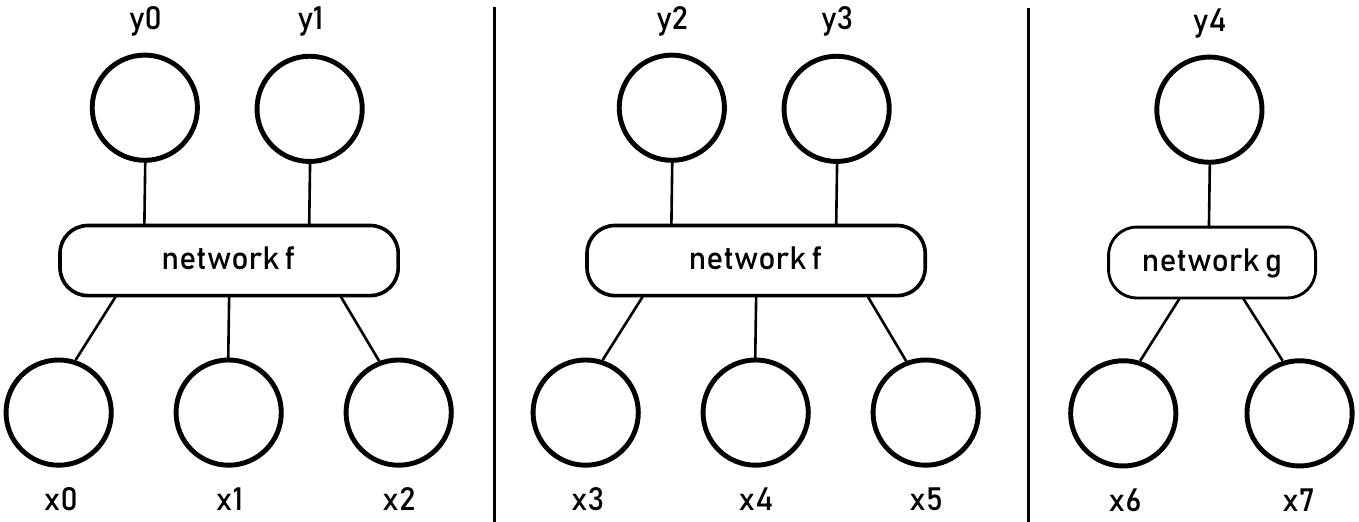}
\caption{Proposed extension to the Marabou query language to support properties involving multiple networks and multiple applications of the same network. Input and output variables are labelled sequentially in the order that the networks are passed to Marabou. The diagram shows the proposed labelling of input and output variables for a property that applies network $f$ to two different inputs and network $g$ to one input.}
\label{fig:meta-network}
\end{figure}

Conceptually the idea is very simple: if a property involves multiple neural network applications, then the set of applications is assigned an order, and additional input and output variables are then assigned sequentially using this order.  An example illustration can be seen in Figure~\ref{fig:meta-network}. In theory this can be seen as composing the networks in parallel, although in practice the different networks will still be stored in individual ONNX files.

We should stress that this extension is not yet implemented by Marabou, although we hope to do so in the near future. Therefore only Vehicle properties that involve a single application of a single network can currently be verified by Marabou.

\section{Vehicle}
\label{sec:language}

\subsection{Specification language}

The Vehicle specification language is a functional language with Haskell-like syntax. At its centre is a small dependently-typed core, upon which various built-in operators and types are then added. Figure~\ref{fig:vehicle-controller-spec} shows one possible formulation  of the specification for the running example, and will be used to explain the key features of the language. The full BNF grammar of the language can be found online.

\begin{figure}[t]
\begin{lstlisting}[language=Haskell]
type InputVector = Tensor Rat [2]

network controller : InputVector -> Rat

currentPosition : InputVector -> Rat
currentPosition x = x ! 0

previousPosition : InputVector -> Rat
previousPosition x = x ! 1

safeInput : InputVector -> Bool
safeInput x = -3.25 <= currentPosition  x <= 3.25 and 
              -3.25 <= previousPosition x <= 3.25

safeOutput : InputVector -> Bool
safeOutput x = let y = controller x in
  -1.25 < y + 2 * currentPosition x - previousPosition x < 1.25

safe : Prop
safe = forall x . safeInput x => safeOutput x
\end{lstlisting}
\caption{The specification of the safety property expressed in Vehicle for car's neural network controller.}
\label{fig:vehicle-controller-spec}
\end{figure}

In order to better abstract away the representation of the inputs of our network, the first line of the specification declares \vehicle{InputVector} to be a synonym for the type of 1-dimensional rational tensors of length 2. Vehicle has a set of builtin types that includes \vehicle{Bool}, \vehicle{Int}, \vehicle{Rat}, \vehicle{Real} and \vehicle{Tensor}. An observant reader may note that neural networks use floating point arithmetic whereas we are using rationals in our specification. We acknowledge this compromises soundness, and aim for Vehicle to support floating point types in the future. Some neural network verifiers have recently been found to have similar unsoundness problems~\cite{jia2021exploiting}.

Next, the car's controller is bound to the name \vehicle{controller} using the \vehicle{network} keyword. As previously discussed, we are only required to provide a name and a type for the network in the specification, and the implementation of the network is provided later to the Vehicle compiler in the form of an ONNX file. Consequently the network remains a black-box function from the perspective of the specification, while still allowing us to write expressive properties which can be statically type-checked.

The local functions \vehicle{currentPosition} and \vehicle{previousPosition} assign meaningful names to the first and second components of the input vector. The \vehicle{!} operator looks up the value of the tensor at the provided index. The \vehicle{safeInput} and \vehicle{safeOutput} declarations then use these to provide human-readable statements of the pre-condition and post-conditions of \AgdaFunction{controller-lemma}.

Finally, the \vehicle{safe} declaration assembles these pieces together to complete the specification. It uses the universal quantifier \vehicle{forall} to bind a new variable \vehicle{x} representing an arbitrary input to the network and then states that whenever \vehicle{safeInput x} is true then \vehicle{safeOutput x} is true as well. Note that the type of \vehicle{safe} is \vehicle{Prop} rather than \vehicle{Bool}. The \vehicle{Prop} type represents the type of boolean expressions whose value cannot be decided within Vehicle itself. Most of the built-ins that use booleans are polymorphic with respect to either \vehicle{Bool} and \vehicle{Prop}. The exceptions are the quantifiers \vehicle{forall} and \vehicle{exists} which always return \vehicle{Prop}, and \vehicle{if then else} which always requires that the condition must be of type \vehicle{Bool}.

\subsection{Compilation to Marabou}
\label{sec:compilation}

The Vehicle compiler is implemented in Haskell and uses a lexer and parser generated by BNFC~\cite{forsberg2004bnf}. The type-checking algorithm is based on the one presented in~\cite{loh2010tutorial}, with the addition of type classes and unification-based term inference.

We will now describe the compiler passes that translate a type-checked Vehicle program into verification queries suitable for Marabou. Note that only the very last pass in the pipeline does anything that is specific to Marabou, and therefore it should be relatively easy to target further verifiers.

\subsubsection{Network type analysis}

The first step is to check the networks declared in the specification against their implementations in the ONNX files provided by the user. Using a custom-written Haskell bindings for the C implementation of ONNX, Vehicle reads the type information from the ONNX file and checks that it matches that declared by the user.

Next it checks that the network type is supported by Vehicle. Although the ONNX format is significantly more expressive, supporting multiple tensor inputs with different sizes and element types, at the moment Vehicle supports only networks of following type, where \vehicle{A}, \vehicle{B} are one of \vehicle{Nat}/\vehicle{Int}/\vehicle{Rat}/\vehicle{Real} and \vehicle{m} and \vehicle{n} are literals:
\begin{center}
\begin{tabular}{c}
\begin{lstlisting}[linewidth=5.8cm]
Tensor A [m] -> Tensor B [n]
\end{lstlisting}
\end{tabular}
\end{center}
However it also allows syntactic sugar for this pattern in the form of:
\begin{center}
\begin{tabular}{c}
\begin{lstlisting}[linewidth=3.8cm]
A -> ... -> A -> B
\end{lstlisting}
\end{tabular}
\end{center}
If the type is in this form, then during this pass, it is normalised to \vehicle{Tensor A [n] -> Tensor B [1]}. This necessitates traversing the program to find any applications of the network, \vehicle{f x1 ... xn} and replacing them with \vehicle{f [x1, ... , xn] ! 0}, where the \vehicle{[ ]} syntax constructs a tensor from the comma-separated list of elements contained within the brackets.

Finally, the network declarations are removed from the program, and the names and types of the networks are stored in the \emph{network context} which is passed along to subsequent stages in the compiler.

\subsubsection{Normalisation} 

The next step is normalisation. As well as performing the standard operations such as beta-reduction, normalisation of builtin operations applied to constants and substituting through any references to top-level functions, some domain-specific operations are required.

Firstly, the verifier input format described in Section~\ref{sec:verifiers}, assigns variables to each individual input and therefore quantifiers over tensor variables must be converted to multiple quantifiers over its elements, e.g. \vehicle{forall (x : Tensor A [2, 2])} is normalised to \vehicle{forall (x11 x12 x21 x22 : A)}. Secondly, after normalisation, only top-level declarations with type \vehicle{Prop} are of interest, as the rest should have been substituted through. Declarations which do not have type \vehicle{Prop} are therefore removed from the program. After normalisation, we are therefore left with the following Vehicle program:
\begin{center}
\begin{tabular}{c}
\begin{lstlisting}[linewidth=10.3cm]
safe : Prop
safe : forall (p0 p1 : Rat) . 
  (3.25 <= p0 <= -3.25) and (3.25 <= p1 <= -3.25) => 
  -1.25 < controller [p0, p1] + 2 * p0 - p1 < 1.25
\end{lstlisting}
\end{tabular}
\end{center}

\subsubsection{Subdivision of queries}
\label{sec:query-subdivision}

Neural network verifiers only support solving existential queries involving conjunctions of numeric equalities and inequalities. We now describe how a Vehicle property is reduced to a set of such queries.

Initially, Vehicle traverses the property making note of the set of quantifiers used. If only existential quantifiers are used then the property passes through this stage untouched. If only universal quantifiers are used then the property is negated. If both types of quantifier are used then the compiler will emit an error.

Next, any \vehicle{if} statements contained within the property are eliminated, using the transformation:
\begin{center}
\begin{tabular}{ccc}
\begin{lstlisting}[linewidth=3.7cm]
if a then b else c
\end{lstlisting} & 
$\quad \Rightarrow \quad$ &
\begin{lstlisting}[linewidth=4.2cm]
a => b and not a => c
\end{lstlisting}
\end{tabular}
\end{center}
Note that this transformation is only valid if the arguments of \emph{if} statement have type \vehicle{Prop} or \vehicle{Bool}. However this may not be the case, e.g. \vehicle{exists x . (if a then x else x + 2) >= 8}. Nonetheless, as the overall type of the Vehicle property is guaranteed to be \vehicle{Prop}, it is always possible to lift the \vehicle{if} expression recursively until the arguments have type \vehicle{Prop}, e.g. \vehicle{exists x . if a then x >= 8 else x + 2 >= 8}, and then perform the elimination.

Next, the expression is converted to disjunctive normal form, with implications being converted to \vehicle{or}s and \vehicle{or} expressions being lifted to the top-level. At this point the following invariants should hold of the property: only existential quantifiers, no negations, no if statements, and the only tensor literals present should be the input to the networks. Our running example is therefore:
\begin{center}
\begin{tabular}{c}
\begin{lstlisting}[linewidth=10.6cm]
exists (p0 p1 : Rat) . 
  (3.25 <= p0 <= -3.25) and (3.25 <= p1 <= -3.25) and
  controller [p0, p1] + 2 * p0 - p1 < -1.25
or 
exists (p0 p1 : Rat) . 
  (3.25 <= p0 <= -3.25) and (3.25 <= p1 <= -3.25) and
  controller [p0, p1] + 2 * p0 - p1 > 1.25
\end{lstlisting}
\end{tabular}
\end{center}

Each disjunction is now split up into its own query. From this point onwards, we will only follow the compilation of the first query in the running example.

\subsubsection{Moving from a functional to relational model of networks}
\label{sec:functional-to-relational}

The next stage is to move from our model of the network as a function to the relational model used by the verifiers. As discussed in Section~\ref{sec:marabou-extension}, we aim to support multiple neural networks applications, and therefore this is a little more involved than one might at first suspect.

The first step is to construct a list of the network applications in the query. Despite supporting multiple applications of the same network, we must be careful not to duplicate applications of the same network to the same input, as doing so would result in an exponential decrease in performance during verification. Therefore in order to avoid this, we first perform a common-sub-expression elimination pass, binding all network applications to fresh variables using let-expressions. We use an efficient hash-based approach procedure using co-de-Bruijn indices~\cite{mcbride2018codebruijn}. Once this pass is complete, we can generate the required list of network applications simply by calculating the set of free variables that occur in the query.

In the next step, we recurse downwards into the expression, keeping track of our position in the list of network applications and replacing every let-bound neural network application with an expression that a) equates the inputs to the application with the input variables, and b) substitutes a list of the output variables for the bound variable in the body of the let-binding. For example, if we are looking at an application of network \vehicle{f} with $n$ inputs and $m$ outputs, and the list of network applications that we have traversed so far has $k$ inputs and $l$ outputs in total then the following transformation would be performed:
\begin{center}
\begin{tabular}{c}
\begin{lstlisting}[linewidth=5.7cm]
let y = f [e1, ..., en] in e
\end{lstlisting} \\ 
$\quad \Rightarrow \quad$ \\
\begin{lstlisting}[linewidth=13.5cm]
[e1, ..., en] == [X<k+1> ... X<k+n>] and e{y/[Y<l+1>, ..., Y<l+m>]}
\end{lstlisting}
\end{tabular}
\end{center}
As with if-elimination in Section~\ref{sec:query-subdivision}, this is only a valid transformation if the type of \vehicle{e} is \vehicle{Prop}. However, this is guaranteed as the common sub-expression elimination inserts the \vehicle{let}s at the top-most position, i.e. just before the quantifiers of the variables bound in the expression.

However, there is one more niggle, as it is necessary to eliminate the variables quantified over by the user. Therefore, when performing the above substitution the compiler tracks which user variables are equated to which introduced input and output variables. Upon recursing back up the tree, when a quantifier is reached, it checks for an equated input or output variable. If there is, we remove the quantifier and substitute the latter through in its place. If there is not, then at the moment, the compiler errors. Note that in future it should be possible to refine this analysis, to reduce the number of such errors. For example the property \vehicle{exists v . f (v + 2) <= 0} would generate the constraint $x0 = v + 2$ which currently errors, but could in fact be rearranged to the form $v = x0 - 2$ and then substituted through.

Finally, the compiler then reruns the normalisation pass in order to simplify the introduced tensor expressions. This leaves the first query from the running example as:
\begin{center}
\begin{tabular}{c}
\begin{lstlisting}[linewidth=5.2cm]
(3.25 <= x0 <= -3.25) and 
(3.25 <= x1 <= -3.25) and
y0 + 2 * x0 - x1 < -1.25
\end{lstlisting}
\end{tabular}
\end{center}

\subsubsection{Conversion to Marabou syntax}

Finally we now specifically target Marabou's query language, recursing through the query and splitting conjunctions into separate assertions. Each individual assertion is first checked for linearity and then rearranged to put the constant on the right hand side of the relation. The left-hand side is then transformed into the required syntax, resulting in the query in Figure~\ref{fig:marabou-queries}a. This query is then written to a user-defined location ready to be fed into Marabou. Marabou proves both queries in approximately 20 seconds on a mid-range laptop.

\subsection{Compilation to Agda}

After compiling the Vehicle specification to Marabou and verifying the resulting queries, we can now use it to complete the overall proof that the car never leaves the road by compiling it to Agda. As before, the Vehicle program is type-checked and the network types compared against those in the ONNX file. The subsequent transformation to Agda code is relatively simple. The only non-trivial insight needed is that any Vehicle expression of type \vehicle{Prop} must be lifted to the \AgdaFunction{Set} type in Agda. This leads to having two different methods of compiling each built-in, depending on whether it is being instantiated with the \vehicle{Bool} or \vehicle{Prop} type. The output of compiling the example specification is:

\begin{code}%
\>[0]\AgdaKeyword{module}\AgdaSpace{}%
\AgdaFunction{ControllerSpec}\AgdaSpace{}\AgdaKeyword{where}\<%
\\
\\
\>[0]\AgdaFunction{InputVector}\AgdaSpace{}%
\AgdaSymbol{:}\AgdaSpace{}%
\AgdaPrimitive{Set}\<%
\\
\>[0]\AgdaFunction{InputVector}\AgdaSpace{}%
\AgdaSymbol{=}\AgdaSpace{}%
\AgdaFunction{Tensor}\AgdaSpace{}%
\AgdaRecord{ℚ}\AgdaSpace{}%
\AgdaSymbol{(}\AgdaNumber{2}\AgdaSpace{}%
\AgdaOperator{\AgdaInductiveConstructor{∷}}\AgdaSpace{}%
\AgdaInductiveConstructor{[]}\AgdaSymbol{)}\<%
\\
\\
\>[0]\AgdaKeyword{postulate}\AgdaSpace{}\AgdaFunction{controller}\AgdaSpace{}%
\AgdaSymbol{:}\AgdaSpace{}%
\AgdaFunction{InputVector}\AgdaSpace{}%
\AgdaSymbol{→}\AgdaSpace{}%
\AgdaRecord{ℚ}\<%
\\
%
%
\\
\>[0]\AgdaFunction{currentPositon}\AgdaSpace{}%
\AgdaSymbol{:}\AgdaSpace{}%
\AgdaFunction{InputVector}\AgdaSpace{}%
\AgdaSymbol{→}\AgdaSpace{}%
\AgdaRecord{ℚ}\<%
\\
\>[0]\AgdaFunction{currentPositon}\AgdaSpace{}%
\AgdaBound{x}\AgdaSpace{}%
\AgdaSymbol{=}\AgdaSpace{}%
\AgdaBound{x}\AgdaSpace{}%
\AgdaSymbol{(}\AgdaOperator{\AgdaFunction{\#}}\AgdaSpace{}%
\AgdaNumber{0}\AgdaSymbol{)}\<%
\\
\\
\>[0]\AgdaFunction{previousPositon}\AgdaSpace{}%
\AgdaSymbol{:}\AgdaSpace{}%
\AgdaFunction{InputVector}\AgdaSpace{}%
\AgdaSymbol{→}\AgdaSpace{}%
\AgdaRecord{ℚ}\<%
\\
\>[0]\AgdaFunction{previousPositon}\AgdaSpace{}%
\AgdaBound{x}\AgdaSpace{}%
\AgdaSymbol{=}\AgdaSpace{}%
\AgdaBound{x}\AgdaSpace{}%
\AgdaSymbol{(}\AgdaOperator{\AgdaFunction{\#}}\AgdaSpace{}%
\AgdaNumber{1}\AgdaSymbol{)}\<%
\\
\\
\>[0]\AgdaFunction{SafeInput}\AgdaSpace{}%
\AgdaSymbol{:}\AgdaSpace{}%
\AgdaFunction{InputVector}\AgdaSpace{}%
\AgdaSymbol{→}\AgdaSpace{}%
\AgdaPrimitive{Set}\<%
\\
\>[0]\AgdaFunction{SafeInput}\AgdaSpace{}%
\AgdaBound{x}\AgdaSpace{}%
\>[12]\AgdaSymbol{=}\AgdaSpace{}%
\AgdaSymbol{(}\AgdaOperator{\AgdaFunction{ℚ.-}}\AgdaSpace{}%
\AgdaSymbol{(}\AgdaOperator{\AgdaInductiveConstructor{ℤ.+}}\AgdaSpace{}%
\AgdaNumber{13}\AgdaSpace{}%
\AgdaOperator{\AgdaFunction{ℚ./}}\AgdaSpace{}%
\AgdaNumber{4}\AgdaSymbol{)}\AgdaSpace{}%
\AgdaOperator{\AgdaDatatype{ℚ.≤}}\AgdaSpace{}%
\AgdaFunction{currentPositon}\AgdaSpace{}%
\AgdaBound{x}\AgdaSpace{}%
\AgdaOperator{\AgdaFunction{×}}\AgdaSpace{}%
\AgdaFunction{currentPositon}\AgdaSpace{}%
\AgdaBound{x}\AgdaSpace{}%
\AgdaOperator{\AgdaDatatype{ℚ.≤}}\AgdaSpace{}%
\AgdaOperator{\AgdaInductiveConstructor{ℤ.+}}\AgdaSpace{}%
\AgdaNumber{13}\AgdaSpace{}%
\AgdaOperator{\AgdaFunction{ℚ./}}\AgdaSpace{}%
\AgdaNumber{4}\AgdaSymbol{)}\AgdaSpace{}\<%
\\
\>[12]\AgdaOperator{\AgdaFunction{×}}\AgdaSpace{}%
\AgdaSymbol{(}\AgdaOperator{\AgdaFunction{ℚ.-}}\AgdaSpace{}%
\AgdaSymbol{(}\AgdaOperator{\AgdaInductiveConstructor{ℤ.+}}\AgdaSpace{}%
\AgdaNumber{13}\AgdaSpace{}%
\AgdaOperator{\AgdaFunction{ℚ./}}\AgdaSpace{}%
\AgdaNumber{4}\AgdaSymbol{)}\AgdaSpace{}%
\AgdaOperator{\AgdaDatatype{ℚ.≤}}\AgdaSpace{}%
\AgdaFunction{previousPositon}\AgdaSpace{}%
\AgdaBound{x}\AgdaSpace{}%
\AgdaOperator{\AgdaFunction{×}}\AgdaSpace{}%
\AgdaFunction{previousPositon}\AgdaSpace{}%
\AgdaBound{x}\AgdaSpace{}%
\AgdaOperator{\AgdaDatatype{ℚ.≤}}\AgdaSpace{}%
\AgdaOperator{\AgdaInductiveConstructor{ℤ.+}}\AgdaSpace{}%
\AgdaNumber{13}\AgdaSpace{}%
\AgdaOperator{\AgdaFunction{ℚ./}}\AgdaSpace{}%
\AgdaNumber{4}\AgdaSymbol{)}\<%
\\
\\
\>[0]\AgdaFunction{SafeOutput}\AgdaSpace{}%
\AgdaSymbol{:}\AgdaSpace{}%
\AgdaFunction{InputVector}\AgdaSpace{}%
\AgdaSymbol{→}\AgdaSpace{}%
\AgdaPrimitive{Set}\<%
\\
\>[0]\AgdaFunction{SafeOutput}\AgdaSpace{}%
\AgdaBound{x}\AgdaSpace{}%
\>[12]\AgdaSymbol{=}\AgdaSpace{}\<%
\\
\>[2]\AgdaOperator{\AgdaFunction{ℚ.-}}\AgdaSpace{}%
\AgdaSymbol{(}\AgdaOperator{\AgdaInductiveConstructor{ℤ.+}}\AgdaSpace{}%
\AgdaNumber{5}\AgdaSpace{}%
\AgdaOperator{\AgdaFunction{ℚ./}}\AgdaSpace{}%
\AgdaNumber{4}\AgdaSymbol{)}\AgdaSpace{}%
\AgdaOperator{\AgdaDatatype{ℚ.<}}\AgdaSpace{}%
\AgdaSymbol{(}\AgdaFunction{controller}\AgdaSpace{}%
\AgdaBound{x}\AgdaSpace{}%
\AgdaOperator{\AgdaFunction{ℚ.+}}\AgdaSpace{}%
\AgdaSymbol{(}\AgdaOperator{\AgdaInductiveConstructor{ℤ.+}}\AgdaSpace{}%
\AgdaNumber{2}\AgdaSpace{}%
\AgdaOperator{\AgdaFunction{ℚ./}}\AgdaSpace{}%
\AgdaNumber{1}\AgdaSymbol{)}\AgdaSpace{}%
\AgdaOperator{\AgdaFunction{ℚ.*}}\AgdaSpace{}%
\AgdaFunction{currentPositon}\AgdaSpace{}%
\AgdaBound{x}\AgdaSymbol{)}\AgdaSpace{}%
\AgdaOperator{\AgdaFunction{ℚ.-}}\AgdaSpace{}%
\AgdaFunction{previousPositon}\AgdaSpace{}%
\AgdaBound{x}\AgdaSpace{}\<%
\\
\>[2]\AgdaOperator{\AgdaFunction{×}}\AgdaSpace{}%
\AgdaSymbol{(}\AgdaFunction{controller}\AgdaSpace{}%
\AgdaBound{x}\AgdaSpace{}%
\AgdaOperator{\AgdaFunction{ℚ.+}}\AgdaSpace{}%
\AgdaSymbol{(}\AgdaOperator{\AgdaInductiveConstructor{ℤ.+}}\AgdaSpace{}%
\AgdaNumber{2}\AgdaSpace{}%
\AgdaOperator{\AgdaFunction{ℚ./}}\AgdaSpace{}%
\AgdaNumber{1}\AgdaSymbol{)}\AgdaSpace{}%
\AgdaOperator{\AgdaFunction{ℚ.*}}\AgdaSpace{}%
\AgdaFunction{currentPositon}\AgdaSpace{}%
\AgdaBound{x}\AgdaSymbol{)}\AgdaSpace{}%
\AgdaOperator{\AgdaFunction{ℚ.-}}\AgdaSpace{}%
\AgdaFunction{previousPositon}\AgdaSpace{}%
\AgdaBound{x}\AgdaSpace{}%
\AgdaOperator{\AgdaDatatype{ℚ.<}}\AgdaSpace{}%
\AgdaOperator{\AgdaInductiveConstructor{ℤ.+}}\AgdaSpace{}%
\AgdaNumber{5}\AgdaSpace{}%
\AgdaOperator{\AgdaFunction{ℚ./}}\AgdaSpace{}%
\AgdaNumber{4}\<%
\\
\\
\>[0]\AgdaKeyword{abstract}\<%
\\
\>[0][@{}l@{\AgdaIndent{0}}]%
\>[2]\AgdaFunction{safe}\AgdaSpace{}%
\AgdaSymbol{:}\AgdaSpace{}%
\AgdaSymbol{∀}\AgdaSpace{}%
\AgdaSymbol{(}\AgdaBound{x}\AgdaSpace{}%
\AgdaSymbol{:}\AgdaSpace{}%
\AgdaFunction{Tensor}\AgdaSpace{}%
\AgdaRecord{ℚ}\AgdaSpace{}%
\AgdaSymbol{(}\AgdaNumber{2}\AgdaSpace{}%
\AgdaOperator{\AgdaInductiveConstructor{∷}}\AgdaSpace{}%
\AgdaInductiveConstructor{[]}\AgdaSymbol{))}\AgdaSpace{}%
\AgdaSymbol{→}\AgdaSpace{}%
\AgdaFunction{SafeInput}\AgdaSpace{}%
\AgdaBound{x}\AgdaSpace{}%
\AgdaSymbol{→}\AgdaSpace{}%
\AgdaFunction{SafeOutput}\AgdaSpace{}%
\AgdaBound{x}\<%
\\
\>[2]\AgdaFunction{safe}\AgdaSpace{}%
\AgdaSymbol{=}\AgdaSpace{}%
\AgdaMacro{checkVehicleProperty}\AgdaSpace{}%
\AgdaKeyword{record}\<%
\\
\>[2][@{}l@{\AgdaIndent{0}}]%
\>[4]\AgdaSymbol{\{}\AgdaSpace{}%
\AgdaField{propertyFile}%
\>[19]\AgdaSymbol{=}\AgdaSpace{}%
\AgdaString{"path/to/property/file.vclp"}\<%
\\
\>[4]\AgdaSymbol{;}\AgdaSpace{}%
\AgdaField{propertyName}\AgdaSpace{}%
\AgdaSymbol{=}\AgdaSpace{}%
\AgdaString{"safe"}\<%
\\
\>[4]\AgdaSymbol{\}}\<%
\end{code}

There are a few things to note. Firstly, the network is declared as a \AgdaKeyword{postulate} and therefore cannot be evaluated within Agda. This is not a fundamental limitation, and with a bit of effort the Haskell bindings created for ONNX could be lifted to Agda. Secondly, the desired proof \AgdaFunction{safe} is within an \AgdaKeyword{abstract} block which prevents code that uses the generated module from depending on the implementation of the proof.

Finally, the definition of the proof is implemented via a macro \AgdaMacro{checkVehicleProperty} which calls out to the Vehicle compiler. A naive implementation would have Agda use a reference to the query files to call Marabou directly. However, while Marabou is running the user would be unable to interact with the file. As Marabou takes over 20 seconds to verify these queries, and potentially much longer for more complex queries, this would unacceptably degrade the user experience. 

Instead we require users to explicitly ask Vehicle to use Marabou to verify the specification. If successful then Vehicle will write out a Vehicle proof file (.vclp) which contains locations and hashes of the ONNX files and the verification status of the specification. It is a reference to this file rather than the original source code that the Agda macro passes back to Vehicle. Vehicle reads this file, uses the hashes to check that the network has not changed on disk, and then returns the verification status of the property to Agda. There is however one missing piece of the puzzle, namely the integrity of the generated Agda code. In theory a malicious or careless user could change the type of the proof in the Agda code which would currently not be detectable by Vehicle. While Vehicle does store the hash of the generated Agda code in the proof file, Agda macros cannot query the location of the source file from which they are called and so Vehicle cannot locate the current Agda file to re-hash. Hopefully this short-coming in Agda can be fixed in the future.

Leaving aside this issue, in summary the generated Agda module is now linked to both the underlying ONNX file and the result of the Marabou verifier. The module can now be imported into the main development outlined in Section~\ref{sec:example} and the \AgdaFunction{safe} proof can now be used in the proof of \AgdaFunction{controller-lemma}. This therefore completes the formal proof of Theorem~\ref{thm:on-road} that, subject to the assumptions that the sensor error is never more than 0.25 and the wind never shifts by more than 1 per unit time, the neural network controlled vehicle will never leave the road.

\section{Conclusions and future work}
\label{sec:conclusions}

This paper has described how the Vehicle tool can be used to express high-level specifications for neural networks. These can then be compiled to low-level queries for neural network verifiers, and to high-level postulates within an interactive theorem prover, linked by hashes, to provided end-to-end verification of neural-network powered systems. We have demonstrated this process by proving the safety of a system involving a 20,000 node neural network controller. All accompanying code, including Vehicle itself, is available online \cite{vehicle2022}.

We expect that as the field of neural network verification matures, a system like Vehicle that integrates automated and interactive theorem proving will be required to ensure that verified neural network specifications support the specifications of the larger systems that contain them.

Before discussing possible future work, we should emphasise that building Vehicle has been a significant undertaking, as many of the necessary components are still relatively immature. Firstly, in order to allow the Vehicle compiler to read the ONNX files we have had to create Haskell bindings for the C version of the ONNX runtime library. Secondly, the current version of Marabou which is implemented in C++, only supports networks in the ONNX format via its Python bindings. We have therefore had to patch support into Marabou for reading ONNX files natively in C++. This involved using undocumented C++ bindings to the ONNX format to manually traverse and parse the networks into Marabou's internal representation. Therefore, while we believe that neural network verification has a bright future, there is significant work to be done before mature, robust, flexible and interoperable tools can be developed.

\subsection{Improved integration}

One possible future direction is to increase the number of ITPs that Vehicle connects to, and in particular those with better support extracting verified code. While there are no reasons why popular theorem provers such as Coq, Isabelle and Idris cannot be targeted, there are also some more interesting potential targets. Among these is a relatively new ITP called KeYmaera X~\cite{fulton2015keymaeraX} which is based on differential dynamic logic and is specifically designed to verify the safety of real-world controllers using continuous dynamics.
Alternatively, adding the ability to compile to a suitable training framework such as Tensorflow would allow the same specification to be used in training.

\subsection{Language features}

There are many additional features that could be added to the Vehicle language. This includes adding top-level \vehicle{parameter} declarations, which would allow users to pass arbitrary values in at compile time rather than hard-coding them in the property. For example, the epsilon value in specifications of robustness properties~\cite{casadio2021property}, or the training dataset with which the network is robust to. Such a feature would facilitate the feedback loop between the verifier and reinforcement learning algorithms implemented in~\cite{ivanov2020case}.

Another improvement would be for the compiler to read the names of the input and output variables from the ONNX file, and use them to automatically bind functions mapping input and output vectors to their elements. In the running example this would remove the need to explicitly define the functions \vehicle{currentPosition} and \vehicle{previousPosition}.

\subsection{A call to arms}

We would like to end with a call to arms to the neural network verifier community. Unlike SMT solvers which have converged on SMTLib as a unified input format, every neural network verifier currently uses their own incompatible input format. While it is impressive that DNNV~\cite{shriver2021dnnv} unifies 13 different verifiers, it is high-level language and therefore is unsuitable to build other tools such as Vehicle on top of. We argue that in order to have a thriving ecosystem of both backends and frontends, a common \emph{low-level} interface is needed for solvers. There is currently a fledgling proposal out there in the form of VNNLib~\cite{vnnlib2022}, but it needs significantly more work to ensure it is suitably expressive to capture all possible properties of interest.

\bibliography{bibliography.bib}

\end{document}